%% file: main.tex
\documentclass[11pt]{article}

\usepackage[margin=1in]{geometry}
\usepackage{amsmath,amssymb,amsthm}
\usepackage{booktabs}
\usepackage{graphicx}
\usepackage{microtype}
\usepackage{xcolor}
\usepackage[numbers]{natbib}
\usepackage{hyperref}
\hypersetup{colorlinks=true, linkcolor=blue!50!black, citecolor=blue!50!black,
            urlcolor=blue!50!black}

\graphicspath{{../figures/}}

\theoremstyle{plain}
\newtheorem{theorem}{Theorem}
\newtheorem{corollary}{Corollary}
\newtheorem{proposition}{Proposition}
\theoremstyle{definition}
\newtheorem{definition}{Definition}
\newtheorem{assumption}{Assumption}
\theoremstyle{remark}
\newtheorem{remark}{Remark}

\newcommand{\E}{\mathbb{E}}
\newcommand{\Prb}{\mathbb{P}}
\newcommand{\KL}[2]{D\!\left(#1 \,\|\, #2\right)}

\newcommand{\arl}{\mathrm{ARL}_0}
\newcommand{\edd}{\mathrm{EDD}}

\title{Quickest Detection of Hallucination Onset:\\
       Delay Bounds and Learned CUSUM Statistics}
\author{Igor Itkin \\
        Independent Researcher \\
        \texttt{ig.itkin@gmail.com}}
\date{}

\begin{document}
\maketitle

\begin{abstract}
Token-level hallucination detectors are evaluated as classifiers, by AUC over
all tokens, yet a streaming monitor is judged by its reaction time: the number
of tokens that pass between the onset of a hallucination and the alarm. We
formulate hallucination onset detection as a quickest change detection problem.
A first-order Markov model of the latent faithful/hallucinated state, validated
on RAGTruth, places the task inside classical change-point theory and yields
Lorden's lower bound on detection delay: about $1.3$ tokens at a false-alarm
rate of $0.01$. We then show that a causal recurrent labeler acts as a CUSUM
with a learned increment. Among the onsets it catches it detects in
$11$--$13$ tokens, against $31$ for a linear per-token baseline, though at this
false-alarm budget every detector catches under a third of onsets and the
recall-honest delay is $56$--$66$ tokens: low-false-alarm onset detection is
hard. A controlled decomposition attributes the speed advantage mostly to a
better per-token score rather than to temporal accumulation. An information-rate
optimality theorem of Donsker--Varadhan type explains the remaining
order-of-magnitude gap: the learned score realizes only $1/4.5$ of the
divergence the features carry, a deficit that recalibration cannot remove, with
the remainder a finite-horizon effect. Classification metrics conceal this
delay structure; sequential analysis makes it measurable. Code:
\url{https://github.com/YehudaItkin/quickest-hallucination-onset}.
\end{abstract}

\input{introduction}
\input{formulation}
\input{theory}
\input{experiments}
\input{discussion}
\input{conclusion}
\input{limitations}

\bibliographystyle{plainnat}
\bibliography{references}

\clearpage
\appendix
\section*{Appendix}
\input{appendix}
\input{proofs}

\end{document}

%% file: introduction.tex
\section{Introduction}
\label{sec:intro}

Put a hallucination detector in front of a language model that streams tokens to a
user, and one number decides whether it is useful: once the model starts making
things up, how many tokens slip out before the detector raises the alarm? A monitor
that flags a hallucination ten tokens after it began has already let a false claim
reach the reader. Yet the field measures token-level detectors almost exclusively
as classifiers, by area under the ROC curve over all tokens. That score rewards
getting the average token right. It says nothing about how quickly a detector
reacts to the moment that matters, the onset.

Most token-level detectors are trained and evaluated exactly this way:
\citet{liu2022hades} framed token-level reference-free hallucination detection as a
benchmark task, and corpora such as RAGTruth \citep{niu2024ragtruth} now support it.
The work closest to our concern already hints at why reaction time is the right lens. \citet{snel2025first} find that the
first token of a hallucination span is far more detectable than its continuation
tokens, an AUC near $0.8$ against near $0.5$, which is precisely a change-point
statement: the onset carries the signal. The same localization question is emerging
at coarser granularity, where \citet{alvarez2026reasoning} locate the first error
in a reasoning chain as an excursion in hidden-state geometry, and the operational
case for reacting during generation is made by \citet{obeso2025realtime}, who flag
hallucinated entities in real time. None of these poses the streaming problem with
an explicit false-alarm--delay trade-off, which is what we add.

That trade-off has a mature theory. Quickest change detection asks exactly the
streaming question: observations switch from one distribution to another at an
unknown time, and a stopping rule must declare the change as fast as possible while
rarely crying wolf \citep{page1954continuous, lorden1971procedures}. It comes with
lower bounds on detection delay that no detector can beat
\citep{lorden1971procedures, lai1998information}, attained asymptotically by Page's
CUSUM \citep{moustakides1986optimal, pollak1985optimal}, with the modern state of
the field surveyed by \citet{xie2021sequential}. This machinery is standard in
quality control, sensor networks, and finance, but to our knowledge it has not been
connected to hallucination detection, where the analogous question, how long a
generation hallucinates before a monitor reacts, is the operational one.
\citet{xie2026sequentialllm} recently argued that sequential alarms and change-point
detection should become standard tools for monitoring deployed LLMs, including
shifts in hallucination rates across queries; we supply a concrete instance one
level down, inside a single generation, where the change point is the onset of a
hallucinated span. When the pre- and post-change densities are unknown or
high-dimensional, a learned statistic replaces the fixed log-likelihood ratio:
\citet{gong2022nncusum} show that the cross-entropy minimizer of such a neural CUSUM
recovers the log-likelihood ratio, the result we use to read a causal recurrent
labeler as a learned CUSUM.

We treat the onset of hallucination as a change-point and ask three questions:

\begin{itemize}
\item[\textbf{RQ1}] What is the smallest detection delay achievable for
hallucination onset at a fixed false-alarm rate?
\item[\textbf{RQ2}] How close do practical detectors come, and does temporal,
learned structure help over per-token scoring?
\item[\textbf{RQ3}] If a gap to the bound remains, where does it come from?
\end{itemize}

Answering RQ1 needs a model of how the hidden faithful/hallucinated state moves. We
show it is a first-order Markov chain: fitting higher orders is statistically
significant but adds under $0.35\%$ of log-likelihood each, so order one captures
$99.7\%$ of the structure. That assumption places the task inside Lorden's minimax
framework and gives a floor on delay of about $1.3$ tokens at a $1\%$ false-alarm
rate.

For RQ2 we compare detectors at a matched false-alarm budget. A parametric CUSUM
that fits Gaussian densities to the feature stream is far off the floor, at $41$
tokens, because a diagonal Gaussian is the wrong likelihood model in $33$
dimensions. A causal recurrent labeler does much better, and we argue it is a
\emph{learned} CUSUM: its recurrent state accumulates evidence and its log-odds
stand in for the cumulative log-likelihood ratio, with the score function learned
rather than assumed. The detector and its $33$-dimensional feature stream come from
prior work on temporal multi-signal hallucination detection; the closest
architectural neighbor reads a generation's log-probabilities as a time series with
a recurrent network \citep{shapiro2026halt}, though at the response level and from a
single signal. Reading the labeler causally presumes hallucination propagates
forward through autoregressive conditioning, for which \citet{akarlar2026trajectory}
give causal support: injecting a hallucinated state corrupts the continuation far
more often than the reverse repair restores it.

At the same false-alarm rate the learned CUSUM detects in $11$--$13$ tokens among the
onsets it catches, against $31$ for the linear per-token baseline. Yet a controlled
decomposition tempers the
temporal reading: a nonlinear per-token model with no sequence already reaches $18$
tokens, so most of the advantage over the linear baseline is a better score; the
sequential accumulation contributes about a quarter of the reduction (significant
under bootstrap) and the extra causal context is within noise.

For RQ3, a large gap to the $1.3$-token floor remains, and a first-order rate lets
us say where most of it comes from. It is not the architecture. The delay of any
score-based detector is set by the information rate its score realizes, and the
learned score realizes only $1/4.5$ of the divergence the features contain; that
shortfall is invariant to recalibration and close to irreducible for these
features. A further factor of two is finite-horizon: the score is so smoothed in
time that the asymptotic correlation penalty overshoots tenfold, and detection
happens faster than the score mixes. The gap is a feature-discriminability problem
first and a finite-horizon question second, not a depth problem. Low-false-alarm
onset detection is also hard in a way the bound does not capture: at the floor's
operating point recall is near $30\%$, so most onsets go uncaught at their first
token, consistent with the difficulty \citet{snel2025first} report.

Our contributions:
\begin{enumerate}
\item We formalize hallucination onset detection as sequential change-point
detection and validate the first-order Markov structure the formulation rests on
(Section~\ref{sec:formulation}, \ref{sec:markov}).
\item We establish Lorden's minimax delay bound for the task and compute it from
the feature divergence (Section~\ref{sec:lorden}).
\item We show a causal recurrent labeler is a learned CUSUM and, with a nonlinear
per-token baseline, decompose its speedup over a linear detector into a better
score (most of it), sequential accumulation, and context, at a matched false-alarm
rate (Section~\ref{sec:learned-cusum}, \ref{sec:experiments}).
\item We give a first-order delay rate for any score-based detector
(Proposition~\ref{prop:rate}) and use it to attribute the order-of-magnitude gap to
a $4.5\times$ information-rate shortfall (invariant to recalibration) and a
finite-horizon residual, showing the asymptotic correlation correction overshoots
because detection precedes mixing (Sections~\ref{sec:experiments},
\ref{sec:closing}).
\end{enumerate}

%% file: formulation.tex
\section{Hallucination Onset as Sequential Change-Point Detection}
\label{sec:formulation}

A generation is a token sequence $x_1, \dots, x_T$. Each token carries a latent
faithfulness state $Z_t \in \{0, 1\}$, where $Z_t = 1$ marks a hallucinated
token, and a human annotator supplies the labels $y_t \in \{0,1\}$ we treat as
ground truth. A \emph{hallucination span} is a maximal run of consecutive
$y_t = 1$. The quantity a streaming monitor cares about is not whether a token
is hallucinated in isolation but \emph{when the first one arrives}.

\begin{definition}[Onset]
The onset of a generation with at least one hallucinated token is
$\theta = \min\{t : y_t = 1\}$, the start of its first span. A generation with
no hallucination has $\theta = \infty$.
\end{definition}

An online detector reads features causally. At step $t$ it has seen
$X_{1:t} = (X_1, \dots, X_t)$, where $X_t$ is the feature vector extracted for
token $t$ (text statistics, NLI signals, generator log-probabilities), and it
must decide whether to raise an alarm using only the past. Formally the detector
is a stopping time $\tau$ with respect to the filtration $\mathcal{F}_t =
\sigma(X_{1:t})$: the event $\{\tau \le t\}$ is determined by $X_{1:t}$ alone.

This is the standard quickest-change-detection setup \citep{xie2021sequential}.
Before the onset, features are drawn from a pre-change law $P_0$ (the
distribution of faithful-token features); from the onset onward they follow a
post-change law $P_1$ (hallucinated-token features). A good detector fires soon
after $\theta$ without firing before it. The two errors trade off, and the trade-off
is measured by two quantities.

\begin{definition}[Operating characteristics]
For a stopping rule $\tau$,
\[
  \arl(\tau) = \E_{\infty}[\tau]
  \qquad\text{and}\qquad
  \edd(\tau) = \E_{\theta}\big[(\tau - \theta)^+\big],
\]
where $\E_\infty$ is taken under the no-change law (every token faithful) and
$\E_\theta$ under a change at $\theta$. The average run length to false alarm
$\arl$ counts the mean number of faithful tokens a detector survives before a
spurious alarm; the expected detection delay $\edd$ counts tokens between onset
and alarm.
\end{definition}

We estimate $\arl$ on the stream of faithful tokens (concatenating the
hallucination-free generations and counting alarms), so it is not capped by the
length of any single document (Appendix~\ref{app:arl0}). This matters. A tempting alternative is a
\emph{per-document} false-alarm rate, the fraction of clean generations that
trigger at least once. Yet a generation has $L \approx 120$ tokens and therefore
$\approx L$ independent chances to misfire, so a per-document rate of $0.01$
demands a per-token rate near $10^{-4}$, about $L$ times stricter than the
per-step rate the theory below controls. Reporting against $\arl$ keeps the
empirical operating point in the same units as the bound: $\arl = \gamma$
corresponds to a per-step false-alarm rate $\alpha = 1/\gamma$, and $\gamma = 100$
is the canonical $\alpha = 0.01$.

The detector's objective, then, is to minimize $\edd$ subject to
$\arl \ge \gamma$. Section~\ref{sec:theory} gives the
lowest delay any detector can achieve under that constraint, and the structural
assumption that makes the bound computable. Section~\ref{sec:experiments}
measures how close real detectors get.

%% file: theory.tex
\section{Theory}
\label{sec:theory}

\subsection{A first-order Markov chain is the right model for the label process}
\label{sec:markov}

The change-point formulation needs a model for how the latent state evolves. We
model the label sequence $\{y_t\}$ as a Markov chain and ask what order is
warranted. Fitting orders $1$ through $4$ on the training labels and comparing by
a likelihood-ratio test gives Table~\ref{tab:markov}. Every higher order is
\emph{statistically} significant ($p < 10^{-3}$, the test has enormous power at
this sample size) and \emph{practically} negligible: each added order lifts the
log-likelihood by less than $0.35\%$. A first-order chain captures $99.7\%$ of
the attainable sequential structure.

\begin{table}[t]
\centering
\begin{tabular}{cccc}
\toprule
Order $k$ & Log-likelihood & Parameters & $\Delta$ vs.\ order $k{-}1$ \\
\midrule
1 & $-12{,}128$ & 2 & --- \\
2 & $-12{,}088$ & 4 & $+0.33\%$ \\
3 & $-12{,}052$ & 8 & $+0.30\%$ \\
4 & $-12{,}014$ & 16 & $+0.32\%$ \\
\bottomrule
\end{tabular}
\caption{Markov-order selection on hallucination labels. Higher orders are
significant by the likelihood-ratio test but add under $0.35\%$ of
log-likelihood each. Order one is sufficient.}
\label{tab:markov}
\end{table}

\begin{assumption}[First-order label dynamics]
\label{ass:markov}
The label process is a first-order Markov chain with transition matrix
\[
  P =
  \begin{pmatrix} 1-p & p \\ 1-q & q \end{pmatrix},
  \qquad p = \Prb(y_t{=}1 \mid y_{t-1}{=}0),\;
          q = \Prb(y_t{=}1 \mid y_{t-1}{=}1).
\]
On our data $p = 0.004$ and $q = 0.907$.
\end{assumption}

We fit $p$ and $q$ by maximum likelihood from the annotated labels: every
adjacent pair $(y_{t-1}, y_t)$ over the training generations is counted into a
$2{\times}2$ table whose rows are then normalized. So $p$ is the fraction of
faithful tokens immediately followed by a hallucinated one (the per-token onset
hazard) and $q$ the fraction of hallucinated tokens followed by another
(persistence); the higher orders of Table~\ref{tab:markov} use the same
count-and-normalize estimator with longer contexts. The rates are displayed to
three decimals, so the persistence ratio $q/p > 200$ quoted below reflects the
unrounded $p \approx 0.0044$ rather than $0.907 / 0.004$; Appendix~\ref{app:data} gives the split sizes and the count table.

Two consequences matter. The onset hazard $p$ is small, so onsets are rare and a
single change-point per generation is the right picture. The persistence $q$ is
large, so once the chain enters the hallucinated state it tends to stay: spans
are geometric with mean length $1/(1-q) \approx 11$ tokens, long enough that the
post-change regime is effectively stationary and the asymptotic theory of the
next subsection applies. The ratio $q/p > 200$ is what makes onset a
genuine change-point rather than i.i.d.\ noise. It also explains a negative
result we report elsewhere: a self-exciting (Hawkes) model does not beat this
two-parameter chain, because the excitation is one-step, not long-range. The
chain is not a simplification of the dynamics. It is the dynamics.

\subsection{The Lorden bound on detection delay}
\label{sec:lorden}

With a single change from a pre-change law $P_0$ to a post-change law $P_1$,
classical theory gives a floor on delay that no causal detector can beat.

\begin{theorem}[\citealp{lorden1971procedures,lai1998information}]
\label{thm:lorden}
Let $D(P_1 \| P_0)$ be the Kullback--Leibler divergence between the post- and
pre-change feature laws. Then every stopping rule $\tau$ with
$\arl(\tau) \ge \gamma$ satisfies
\[
  \edd(\tau) \;\ge\; \frac{\ln \gamma}{\KL{P_1}{P_0}}\,\bigl(1 + o(1)\bigr)
  \qquad (\gamma \to \infty),
\]
and the CUSUM rule of \citet{page1954continuous} attains this floor
asymptotically \citep{moustakides1986optimal}.
\end{theorem}

The bound is intuitive: each post-change observation supplies on average
$D(P_1\|P_0)$ nats of evidence, a false-alarm budget $\gamma$ requires
accumulating $\ln\gamma$ nats before stopping, so the fastest possible detector
needs $\ln\gamma / D(P_1\|P_0)$ observations. Estimating the feature divergence
on our $33$-dimensional signal under a diagonal-Gaussian model gives
$D(P_1\|P_0) \approx 3.5$ nats, hence

\begin{equation}
\label{eq:bound}
  \edd_{\min}(\gamma = 100)
  \;=\; \frac{\ln 100}{3.5}
  \;\approx\; 1.3 \text{ tokens at } \alpha = 0.01 .
\end{equation}

An oracle that observed the labels themselves would do even better: the
label-space divergence is $\approx 4.6$ nats (a floor of $1.0$ token), and
because $q = 0.907$ makes the post-change label nearly deterministic, such an
oracle detects essentially at the onset. The gap between $1.3$ tokens (best
possible from features) and what real feature detectors achieve is the subject of
Section~\ref{sec:experiments}.

\begin{remark}
Equation~\eqref{eq:bound} is a per-step statement: it controls the false-alarm
rate per token, and $\gamma$ is the mean number of tokens between false alarms.
This is why we report $\arl$ rather than a per-document rate
(Section~\ref{sec:formulation}); the two differ by a factor of the document
length.
\end{remark}

\subsection{A causal recurrent labeler is a learned CUSUM}
\label{sec:learned-cusum}

CUSUM, the rule that attains the bound, accumulates the log-likelihood ratio and
resets at zero:
\begin{equation}
\label{eq:cusum}
  S_t = \max\!\Big(0,\; S_{t-1} + \log\frac{p_1(X_t)}{p_0(X_t)}\Big),
  \qquad \tau = \min\{t : S_t \ge h\}.
\end{equation}
It needs the two densities $p_0, p_1$. When they are misspecified (as a
diagonal Gaussian is for a $33$-dimensional feature stream), the increment is the
wrong score and $S_t$ accumulates noise instead of signal.

A causal (forward) recurrent labeler replaces the fixed log-likelihood ratio
with a learned one. It maintains a hidden state $h_t = f_\phi(h_{t-1}, X_t)$ and
emits a posterior $\hat{p}_t = \sigma(w^\top h_t)$, so its log-odds
$\operatorname{logit}\hat{p}_t$ are a learned, nonlinear, accumulated statistic of
the whole causal history $X_{1:t}$. To make the correspondence with
\eqref{eq:cusum} precise, view the generation as a two-state hidden Markov model:
the latent state $Z_t$ follows the chain of Assumption~\ref{ass:markov}, and the
feature vector is drawn from a state-conditional emission
$X_t \mid Z_t = z \sim p_z$.

\begin{assumption}[Emission regularity]
\label{ass:emission}
The emissions $p_0, p_1$ have bounded, Lipschitz densities on the feature space,
with finite divergence $\KL{p_1}{p_0}$.
\end{assumption}

\begin{corollary}[Optimal onset detection is a thresholded filter, realizable by a recurrent network]
\label{cor:learned-cusum}
Under Assumptions~\ref{ass:markov}--\ref{ass:emission}:
\begin{enumerate}
\item[(i)] \emph{(Optimality.)} The Bayes-optimal onset detector that minimizes
expected delay at a fixed false-alarm level is a threshold rule on the change
posterior $\pi_t = \Prb(\theta \le t \mid X_{1:t})$, a finite-dimensional filter
that obeys a forward recursion on the two-state belief
\citep{shiryaev1963optimum, ford2023exactly}.
\item[(ii)] \emph{(Score.)} The minimizer of the per-token cross-entropy loss is
the exact log-likelihood ratio $\varphi^\star(x) = \log\!\big(p_1(x)/p_0(x)\big)$
\citep{gong2022nncusum}, which is the CUSUM increment of \eqref{eq:cusum}.
\item[(iii)] \emph{(Realizability.)} For every $\varepsilon > 0$ there is a causal
recurrent network whose output approximates the filter in (i) within
$\varepsilon$, uniformly in $t$ \citep{bishop2023recurrent}.
\end{enumerate}
Chaining (i)--(iii): a causal recurrent labeler trained by cross-entropy is a
consistent estimator of the optimal sequential detector. As its approximation
error (iii) and the finite-sample error of its score (ii) vanish, the $\arl$ and
$\edd$ of its thresholded output approach those of the optimal CUSUM, whose delay
is the floor of Theorem~\ref{thm:lorden}.
\end{corollary}

We state this as a corollary because it assembles three existing results
(Shiryaev optimality for hidden Markov models, the cross-entropy/log-likelihood-ratio
identity, and universal filter approximation) into a statement about hallucination
onset, whose hypotheses Assumptions~\ref{ass:markov}--\ref{ass:emission} verify;
the work is in checking the hypotheses, not in a new proof. It is a
limit statement. To turn it into a rate, read the detector off the posterior in
two ways: threshold $\hat{p}_t$ directly, or feed it through the explicit
accumulation
\begin{equation}
\label{eq:learned-cusum}
  S_t = \max\!\Big(0,\; S_{t-1} + \operatorname{logit}\hat{p}_t - k\Big),
  \qquad k = \tfrac{1}{2}\big(\mu_0 + \mu_1\big),
\end{equation}
where $\mu_0, \mu_1$ are the mean log-odds the model assigns to faithful and
hallucinated tokens. The reference value $k$ is the textbook CUSUM choice that
centers a faithful token below zero and a hallucinated token above it, whatever
the raw posterior's calibration.

\begin{proposition}[First-order delay of a general-score CUSUM]
\label{prop:rate}
Run \eqref{eq:learned-cusum} on increments $Y_t = \operatorname{logit}\hat{p}_t - k$
with $\E_0[Y] < 0 < \E_1[Y]$, and let $\omega > 0$ solve $\E_0[e^{\omega Y}] = 1$.
As the threshold grows, $\arl = e^{\omega h}(1 + o(1))$ and
$\edd = h / \E_1[Y]\,(1 + o(1))$ \citep{siegmund1985sequential}, so
\[
  \edd \;\approx\; \frac{\ln \arl}{I(\hat{g})},
  \qquad I(\hat{g}) = \omega\,\E_1[Y]
\]
is governed by the \emph{realized information rate} $I(\hat{g})$ of the score
(proof in Appendix~\ref{app:proofs}). The true log-likelihood ratio gives
$\omega = 1$ and $I = \KL{p_1}{p_0}$, recovering Theorem~\ref{thm:lorden}; for any
other score $I(\hat{g}) \le \KL{p_1}{p_0}$ by Theorem~\ref{thm:info-opt} below, so
the multiplicative gap to the floor is exactly $\KL{p_1}{p_0} / I(\hat{g})$.
\end{proposition}

That the log-likelihood-ratio score is delay-optimal is classical
\citep{moustakides1986optimal}; what we add is the \emph{variational form} of the
shortfall. Writing the rate of a general score through the Lundberg exponent makes
$I(s)$ coincide with a Donsker--Varadhan functional, so the gap $D/I(s)$ is exactly
the score's Donsker--Varadhan deficit. This is the one place we give a
self-contained proof, and it is what ties the floor to a quantity we can measure on
the learned score (Section~\ref{sec:experiments}).

\begin{theorem}[Information-rate optimality]
\label{thm:info-opt}
Let $s$ be any bounded score with increment $Y = s(X) - k$ obeying
$\E_0[Y] < 0 < \E_1[Y]$, and let $\omega > 0$ solve $\E_0[e^{\omega Y}] = 1$. Then
the realized information rate of Proposition~\ref{prop:rate} satisfies
\[
  I(s) \;=\; \omega\,\E_1[Y] \;\le\; \KL{p_1}{p_0},
\]
with equality if and only if $s$ is an affine function of the log-likelihood ratio
$\log(p_1/p_0)$ ($P_0$-almost everywhere). Hence the log-likelihood-ratio score is
delay-optimal and attains the floor of Theorem~\ref{thm:lorden}; any other score's
delay exceeds the floor by the factor $\KL{p_1}{p_0}/I(s)$.
\end{theorem}

The proof (Appendix~\ref{app:proofs}) applies the Donsker--Varadhan formula to
$f = \omega Y$, where the Lundberg normalization $\E_0[e^{\omega Y}] = 1$ removes
the log-moment term.

Proposition~\ref{prop:rate} and Theorem~\ref{thm:info-opt} turn ``a gap remains''
into a number we can read off the data, and isolate the leading cause. Two finite-sample effects hold a
trained labeler back, and Section~\ref{sec:experiments} measures both.

\begin{remark}[Where the gap comes from]
\label{rem:gap}
The dominant factor is \emph{the realized divergence}: the learned score realizes
$I(\hat{g}) < \KL{p_1}{p_0}$, and Proposition~\ref{prop:rate} turns that shortfall
directly into delay; \citet{gong2022nncusum} reach the same conclusion through a
maximum-mean-discrepancy surrogate looser than the KL rate. This shortfall is a
property of the score's shape, not its scale, and Section~\ref{sec:closing} shows
it is close to irreducible for our features. A smaller residual is
\emph{finite-horizon}: the score increments are strongly correlated, yet detection
is faster than the score mixes, so the asymptotic dependent-data rate overshoots
rather than tightens the prediction (Section~\ref{sec:closing}).
\end{remark}

This reframes our earlier finding, that a recurrent model beats a linear
per-token classifier by a wide margin, as a statement about the score function of
Corollary~\ref{cor:learned-cusum}(ii). The next section measures, at a matched
false-alarm rate, how much that buys, how much of it is the sequential
accumulation as opposed to a better per-token score, and how far it still sits
from the floor.

%% file: experiments.tex
\section{Experiments}
\label{sec:experiments}

\subsection{Setup}

We evaluate on the RAGTruth test split \citep{niu2024ragtruth}: $2{,}700$
generations, $943$ of them containing at least one hallucination, $1{,}757$
clean. Features are the $33$-dimensional per-token signal of our base system
(text statistics, NLI, generator log-probabilities). Five causal detectors,
matched at a common $\arl$ by sweeping their thresholds:

\begin{itemize}
\item \textbf{Naive Gaussian CUSUM}: the parametric rule \eqref{eq:cusum} with
$p_0, p_1$ fit as diagonal Gaussians on the feature stream. The misspecified
baseline.
\item \textbf{LogReg (per-token)}: a logistic regression posterior thresholded
token by token. Linear, no accumulation.
\item \textbf{HistGBM (per-token)}: a gradient-boosted per-token classifier on the
same features. Nonlinear but still per-token (no accumulation); it isolates how
much of a recurrent model's edge is a better score rather than the sequence.
\item \textbf{ForwardGRU (threshold)}: a forward recurrent labeler whose
posterior is thresholded directly. Its recurrent state already accumulates, so
this is the learned CUSUM read off without an explicit sum.
\item \textbf{ForwardGRU (CUSUM)}: the same posterior fed through the explicit
accumulation \eqref{eq:learned-cusum}.
\end{itemize}

We report two delays. \emph{Delay among detected} is the mean of $\tau - \theta$
over generations the detector catches: its speed when it fires. \emph{Censored
$\edd$} averages over \emph{all} hallucination generations, charging a miss the
maximum possible delay (tokens remaining after the onset); it cannot be inflated
by a low recall. We give recall alongside both.

\subsection{Results}

\begin{table}[t]
\centering
\begin{tabular}{lccc}
\toprule
& \multicolumn{3}{c}{Delay among detected, tokens (recall)} \\
\cmidrule(lr){2-4}
Detector & $\arl = 50$ & $\arl = 100$ & $\arl = 200$ \\
\midrule
LogReg (linear per-token)    & $30.7$ ($0.35$) & $30.8$ ($0.31$) & $35.9$ ($0.20$) \\
HistGBM (nonlinear per-token)& $14.4$ ($0.38$) & $17.9$ ($0.40$) & $21.4$ ($0.33$) \\
Naive Gaussian CUSUM         & \multicolumn{3}{c}{$\approx 41$ (misspecified, off-scale)} \\
ForwardGRU-shuffled (CUSUM)  & $15.3$ ($0.17$) & $15.6$ ($0.21$) & $15.2$ ($0.29$) \\
ForwardGRU (threshold)       & $11.2$ ($0.26$) & $13.4$ ($0.30$) & $14.0$ ($0.27$) \\
ForwardGRU (CUSUM)           & $\mathbf{9.0}$ ($0.19$) & $\mathbf{11.5}$ ($0.24$) & $\mathbf{13.3}$ ($0.32$) \\
\midrule
Lorden bound (features)      & $1.1$ & $1.3$ & $1.5$ \\
Oracle (observes labels)     & \multicolumn{3}{c}{$\approx 0$ (detects at onset)} \\
\bottomrule
\end{tabular}
\caption{Detection delay at matched false-alarm budgets. A \emph{nonlinear}
per-token model (HistGBM) already closes most of the gap between the linear
baseline and the recurrent CUSUM, so the recurrent model's edge is mostly a better
score, not the sequence (decomposition in Figure~\ref{fig:operating}). All stay an
order of magnitude above the floor; recall near $30\%$ reflects the difficulty of
low-false-alarm onset detection.}
\label{tab:main}
\end{table}

\begin{figure}[t]
\centering
\includegraphics[width=0.72\linewidth]{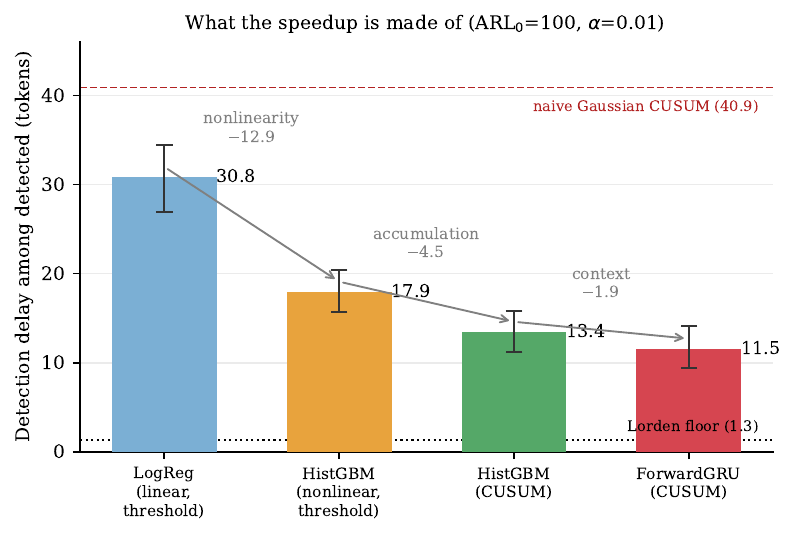}
\caption{Where the speedup comes from, at $\arl = 100$. From the linear per-token
baseline ($30.8$) to the recurrent CUSUM ($11.5$), most of the reduction is the
nonlinear per-token score ($-12.9$, significant); the sequential accumulation
($-4.5$, significant) and causal context ($-1.9$, within noise) are smaller. Error
bars are $95\%$ bootstrap CIs over documents. The naive Gaussian CUSUM (dashed) is
off-scale; all detectors sit far above the Lorden floor (dotted).}
\label{fig:operating}
\end{figure}

Table~\ref{tab:main} and Figure~\ref{fig:operating} give the picture at
$\arl = 100$ (the $\alpha = 0.01$ operating point of the bound).

Most of the recurrent model's speedup over the linear baseline is a better
per-token score, not the sequence.
At $\arl = 100$ the ForwardGRU CUSUM detects in $11.5$ tokens against $30.8$ for
the linear per-token baseline, a $2.7\times$ speedup that is tempting to credit to
the temporal model. Yet a \emph{nonlinear} per-token classifier with no sequence
at all (HistGBM) already detects in $17.9$ tokens, covering most of that ground.
Decomposing the $30.8 \to 11.5$ reduction (Figure~\ref{fig:operating} and Appendix~\ref{app:decomp}; brackets are
$95\%$ bootstrap CIs over documents): the nonlinear score accounts for $-12.9$
tokens $[8.8, 17.0]$ (LogReg to HistGBM), and the CUSUM accumulation for $-4.5$
$[1.8, 7.1]$ (HistGBM threshold to CUSUM), both significant. The further causal
context, $-1.9$ $[-1.0, 4.7]$ (HistGBM-CUSUM to ForwardGRU-CUSUM), is within noise.
About two-thirds of the advantage over a linear detector is a better per-token
score; the sequential accumulation that Corollary~\ref{cor:learned-cusum}
formalizes is real but modest. The accumulation helps the linear and the nonlinear
score about equally ($-4.5$ tokens each), which marks it as a genuine, if
secondary, effect rather than an artifact of the score. A same-architecture control
sharpens the point. A ForwardGRU trained on token-\emph{shuffled} sequences (same
model, no temporal order) detects in $15.6$ tokens, so temporal order is worth about
$4$ tokens to the recurrent model itself. Even so, a nonlinear per-token score with
only CUSUM accumulation (HistGBM-CUSUM, $13.4$) already matches the full recurrent
CUSUM ($11.5$). Order helps the network, but it is not needed to reach its performance.

The parametric baseline fails for a specific, nameable reason.
A diagonal Gaussian is a poor model of the $33$-dimensional feature law, so the
log-likelihood increment in \eqref{eq:cusum} is the wrong score and the statistic
accumulates noise. Its $41$-token delay is worse than even the linear posterior.
Adding more features makes it worse, not better, because each extra dimension
adds Gaussian-model error. The bound is reachable in principle, but only with a
score function close to the true $\log(p_1/p_0)$, which is exactly what the
recurrent model learns and the Gaussian does not.

Most of the remaining gap is a score-shape shortfall.
The ForwardGRU CUSUM detects in $11.5$ tokens, about $9\times$ the $1.3$-token
floor. Proposition~\ref{prop:rate} locates most of it. The learned score realizes
an information rate $I(\hat{g}) = \omega\,\delta_1 = 0.78$ nats per token
($\omega = 0.95$, $\delta_1 = 0.82$), well below the $3.5$ nats of the feature
divergence, so its i.i.d.\ first-order delay is $\ln(100)/I(\hat{g}) = 5.9$
tokens, a factor of $D/I(\hat{g}) = 4.5$ above the floor. The tilting is efficient
($\omega \approx 1$), but the per-token separation is small ($\delta_1 = 0.82$ nats
against the $3.5$ of the feature divergence). By $D/I = 4.5$ the score is far from a
valid likelihood ratio, and the shortfall is in per-token discrimination, not in how
the statistic accumulates.
This $4.5\times$ is a deficit \emph{relative to the i.i.d.\ first-order rate} of
Proposition~\ref{prop:rate}, not of the true gap: the rate's i.i.d.\ hypothesis is
violated by the learned score (Section~\ref{sec:closing} shows its increments are
strongly correlated). So the decomposition that follows is a factorization against
that idealization. The remaining factor of about $2$, from $5.9$ predicted to
$11.5$ observed, is exactly where the i.i.d.\ rate breaks; we examine both next.

\subsection{What moves the realized rate}
\label{sec:closing}

Rescaling the score does not move its rate, and reshaping it barely does.
Temperature scaling $s \mapsto s/T$ leaves $I(\hat{g})$ exactly unchanged
($\omega \mapsto \omega T$, $\delta_1 \mapsto \delta_1/T$): the shortfall is in the
\emph{shape} of the score, not its calibrated scale, so recalibrating confidence
cannot help. A monotone-nonlinear reshaping (isotonic regression of the posterior
to the labels) does change the shape, yet recovers only $+12\%$ of $I(\hat{g})$
($0.78 \to 0.87$). The $4.5\times$ shortfall is close to irreducible for these
features: the per-token signal does not separate faithful from hallucinated tokens
as sharply as the marginal feature divergence suggests. Shrinking it needs more
discriminative features, not a retuned model.

The residual factor of two is finite-horizon, not an asymptotic correlation effect.
The learned score is strongly smoothed in time: its clean-stream autocorrelation
decays slowly ($\rho_1 = 0.94$), with integrated autocorrelation time
$\tau \approx 22$. Correlated increments should, asymptotically, deflate the rate,
and indeed the adjustment coefficient read from the detector's own
$\arl$--threshold curve is $\omega^\star \approx 0.044$, below the marginal
$\omega = 0.95$ by precisely the factor $\tau$. Yet the asymptotic dependent-data
rate \citep{lai1998information} then predicts a delay near $126$ tokens, an order
of magnitude past what we observe. It overshoots because detection here is
\emph{faster than mixing}: the onset is caught in $\sim 11$ tokens, well inside the
score's $\tau \approx 22$ correlation time, and the reset CUSUM floors $\arl$ at
the mean document length. Neither limit is exact in this regime. The i.i.d.\ rate
is a lower bound on delay ($5.9$); the realized $11.5$ sits a factor of $2$ above
it, and pinning that down is a finite-horizon question, not an asymptotic
correlation correction.

Low-false-alarm onset detection is intrinsically hard.
Recall is near $30\%$ at $\arl = 100$ for every drift detector, so the
recall-honest censored $\edd$ is $56$--$66$ tokens: most onsets
are simply not caught within a tight false-alarm budget. This is consistent with
\citet{snel2025first}, who find the first hallucinated token detectable at an AUC
around $0.8$ rather than $1.0$. The first token of a span is the most detectable
one, but it is far from trivially detectable, and a streaming monitor that must
hold its false-alarm rate down will miss most onsets at their first token.

\subsection{Robustness of the deficit to richer features and a rate-aware objective}
\label{sec:robust}

The two prescriptions of Section~\ref{sec:closing} --- more discriminative features, a
better-shaped score --- are testable, and we test both. They fail, and the way they
fail says what the deficit is.

First the anatomy of the rate. For a centered reference the increment has
$\E_0[Y] = -m$ and $\E_1[Y] = +m$ with $m = \tfrac12(\mu_1 - \mu_0)$ the half
mean-gap of the log-odds, and for near-Gaussian increments the adjustment
coefficient is $\omega = 2m/\sigma_0^2$, so the realized rate is
\begin{equation}
\label{eq:rate-anatomy}
  I(\hat{g}) \;=\; \omega\,\delta_1 \;=\; \frac{2m^2}{\sigma_0^2},
\end{equation}
where $\sigma_0$ is the standard deviation of the score increment on clean tokens.
The form holds on our data: the empirical $I$ is within $10$--$25\%$ of
\eqref{eq:rate-anatomy} across LogReg, ForwardGRU and BiGRU, the increments being
close to Gaussian (Appendix~\ref{app:robust}). Equation~\eqref{eq:rate-anatomy}
also factors the deficit: $D/I = (D/D_{\mathrm{inc}})\,(D_{\mathrm{inc}}/I)$, where
$D_{\mathrm{inc}}$ is the divergence of the scalar increment itself. The accumulation
factor $D_{\mathrm{inc}}/I \approx 1.1$ --- the CUSUM is already near-optimal given
the score --- and the whole shortfall is $D/D_{\mathrm{inc}} \approx 4$: the scalar
log-odds is a lossy summary of the $33$-dimensional feature law.

More features do not help, even when they carry genuine new divergence. The natural
black-box candidate is self-consistency: resample the generator $K = 5$ times and
score each token by how much the resamples contradict its sentence
\citep{manakul2023selfcheckgpt}. On RAGTruth this adds $0.83$ nats of real,
non-Gaussian divergence to the feature law (a nearest-neighbour estimate, against a
random-feature control that adds none). Yet a ForwardGRU-CUSUM trained on the
augmented features matches the baseline within seed noise on every axis we
care about --- token AUC, $I(\hat{g})$, and censored $\edd$ (five seeds,
Appendix~\ref{app:robust}). The divergence is in the features; the trained scalar
score does not realize it.

Retuning the objective does not help either, and \eqref{eq:rate-anatomy} says why.
The rate depends only on the standardized separation $m/\sigma_0$, which is
invariant to the scale of the log-odds. Adding a penalty on the clean-token
increment variance $\sigma_0$ to the cross-entropy loss lowers $\sigma_0$ as
intended, but lowers $m$ in step: across penalty strengths $m/\sigma_0$ holds at
$0.57 \pm 0.01$ and the censored $\edd$ does not move ($55.4 \to 54.9$ tokens,
within seed noise). This is the training-time twin of the temperature-scaling
invariance in Section~\ref{sec:closing}: a scale-only intervention, post-hoc or in the
loss, cannot change a rate that is set by the score's shape.

Both routes fail for one reason. The scalar score sits at the ceiling of what a
one-dimensional causal statistic of these features can carry: $m/\sigma_0$ does not
rise when the feature divergence is raised (the score does not pick it up) or when
the objective is retuned (it only rescales). What the factorization isolates is the
$D/D_{\mathrm{inc}}$ bottleneck --- the compression of a high-dimensional feature
law into one log-odds.

A multivariate accumulator of the mean shift is the natural way to lift that
bottleneck, and on these signals it does not. Measured by delay among detected, an
optimal linear readout of the recurrent hidden state (the LDA projection
$\Sigma_h^{-1}(\mu_1-\mu_0)$ accumulated as a CUSUM) matches the trained scalar head
to within a token ($12.7$ against $11.5$). A Hotelling generalized-likelihood-ratio
statistic on the same $64$-dimensional state is worse ($18.1$), defeated by the noise
of a quadratic form in many dimensions (Appendix~\ref{app:robust}). The cross-entropy
head already reads the representation near-optimally for detection.

The bottleneck is also narrower than $D = 3.5$ suggests. That divergence counts
variance and shape differences between the two regimes, but a drift CUSUM exploits
only the mean shift, and on these signals the mean shift is the smaller part: the
raw-feature Mahalanobis separation is $0.40$ nats against the $3.5$ the
diagonal-Gaussian divergence reports, so the regimes differ more in the spread of
the features than in their location.\footnote{A detector built on that second-order
structure recovers part of what the drift score misses. A CUSUM on the covariance term
$\tfrac12 x^\top(\Sigma_0^{-1}-\Sigma_1^{-1})x$ of the full Gaussian log-likelihood ratio
reaches a recall $0.065$ higher than the best per-token scalar (a gradient-boosted
classifier) at $\arl = 100$ (95\% bootstrap CI $[0.03, 0.10]$), at comparable
per-detection delay and only on the open-weight generators. The mean term alone stays
below the scalar, so the gain is the covariance change, not a location shift. The
effect is modest and we do not pursue it here.} The recurrent model concentrates the usable
mean shift into its representation ($1.5$ nats) and the scalar drift captures most of
it. The deficit is a property of the information a causal drift carries, not of the
rule that accumulates it; closing it needs features whose location, not just their
variance, separates the regimes --- internal-state probes among them, at the cost of
black-box applicability.

%% file: discussion.tex
\section{Discussion}
\label{sec:discussion}

Treating hallucination onset as a change-point buys something the classification
view lacks: a yardstick. Token AUC rewards getting the average token right and
says nothing about reaction time, whereas the Lorden floor states, in the units a
deployment cares about, how fast any detector could possibly be. Read against that
floor, our measurements carry two lessons.

The first is that the temporal frame is the right one but is not where the
empirical win comes from. A causal recurrent labeler is a learned CUSUM, and the
learning matters: it detects two to three times faster than a parametric CUSUM or
a linear per-token model at a matched false-alarm rate. What the learning buys,
though, is mostly a better per-token score. A nonlinear per-token classifier with
no sequence already closes most of the gap to the recurrent CUSUM, leaving the
sequential accumulation as a real but secondary effect and the extra causal
context within bootstrap noise. We read this as a clarification of temporal
modeling rather than a demotion of it: the sequential-detection framework is what
makes the delay measurable and gives the recurrent labeler its CUSUM
interpretation, while the magnitude of the empirical speedup is set by score
quality.

The second lesson concerns the order-of-magnitude gap that remains between the
best detector and the floor, and it is not an argument for deeper models. The
delay of any score-based detector is fixed by the information rate its score
realizes, and our learned score realizes only about a fifth of the divergence the
features carry, a deficit that is invariant to recalibration and barely moves
under monotone reshaping. The bound is a property of the information in the
features, so the way to halve the achievable delay is to extract features that
separate the two regimes more sharply, not to add depth. This is a general
statement about score-based onset detection, not a quirk of our architecture.

A third point is easy to miss behind an AUC and worth stating plainly. At the
floor's operating point our detectors catch under a third of onsets at their first
token. Low-false-alarm streaming detection is genuinely hard: a monitor that
promises to flag hallucinations as they happen, without burying the user in false
alarms, will miss most onsets at the first opportunity and catch them a span
later, if at all. That is the regime real deployments live in, and a token-level
AUC hides it. Because the detector needs no access to the generating model, it
remains applicable to closed-source APIs where white-box probes do not; the
sequential view simply makes its true operating cost visible.

%% file: conclusion.tex
\section{Conclusion}
\label{sec:conclusion}

Hallucination onset is a change-point, and treating it as one buys a yardstick the
classification view lacks: a hard floor on how fast any detector can react. For
RAGTruth that floor is about $1.3$ tokens at a $1\%$ false-alarm rate, and it
rests on a fact worth stating plainly, that the faithful/hallucinated state is a
first-order Markov chain and nothing more elaborate is needed. Against that floor,
a causal recurrent labeler acts as a learned CUSUM and detects, among the onsets it
catches, two to three times faster than a parametric CUSUM or a linear per-token
baseline, though a controlled
decomposition shows most of that advantage is a better per-token score rather than
the sequence, and the detector still sits an order of magnitude above the bound.

Two directions follow, and the analysis singles out the first as dominant. The
larger lever is the divergence the score can realize: a feature set that doubles
$D(P_1\|P_0)$ halves the achievable delay, and unlike recalibration it actually
moves the rate. The second is theoretical. The factor-of-two residual lives in a
regime where detection is faster than the score mixes, so neither the i.i.d.\
first-order rate nor the asymptotic correlation correction is tight; a
finite-horizon analysis of CUSUM delay under a strongly autocorrelated score would
close it. Both are concrete, and both are measurable against the bound this paper
puts on the table.

%% file: limitations.tex
\section{Limitations}
\label{sec:limitations}

The bound and the theorem rest on idealizations worth stating plainly, because
each one bears on how the numbers should be read.

The first is that the emissions are not i.i.d.\ given the state.
Theorem~\ref{thm:lorden} and the divergence in \eqref{eq:bound} treat the
per-token features as independent draws from $P_0$ or $P_1$ conditional on the
state, yet several of our $33$ features are windowed or cumulative, so consecutive
tokens are correlated even within one regime. Correlated observations carry less
information per token than i.i.d.\ ones, so the true per-token evidence is below
$D(P_1\|P_0)$ and the floor of $1.3$ tokens is, if anything, optimistic; the gap
we report is therefore a lower bound on the true gap.

The divergence itself is only a diagonal-Gaussian estimate. We compute
$D(P_1\|P_0) \approx 3.5$ nats under a diagonal-Gaussian model of the feature law,
which ignores cross-feature dependence and non-Gaussian shape, and a different
estimator would move the floor. We use this model because the naive CUSUM baseline
assumes it too, which keeps that comparison fair, but the precise value of the
bound should be read as an order of magnitude rather than a constant to three
digits. As a check, a non-parametric $k$-nearest-neighbour estimator
\citep{wang2009divergence} of the same divergence gives $D \approx 2.8$ nats
(floor $\approx 1.6$ tokens), in line with the diagonal-Gaussian value, so the
floor is robust to the estimator.

The delay rate is first-order, and its residual is measured rather than bounded.
Proposition~\ref{prop:rate} expresses the delay through the realized information
rate $I(\hat{g})$, but it is first-order in the threshold and treats the score
increments as i.i.d. Section~\ref{sec:closing} shows the leftover factor of two is
finite-horizon: the asymptotic correlation correction overshoots by an order of
magnitude because detection precedes the score's mixing time, so neither limit is
tight. We measure that residual rather than bound it; a finite-horizon rate for
CUSUM under a strongly autocorrelated score would be stronger, and harder.

Our analysis also covers only the first onset of each generation. The optimality
in Corollary~\ref{cor:learned-cusum}(i) is for a single change-point, so
generations with several spans, and the question of re-arming the detector after a
span ends, are out of scope; the first-passage view matches a deployment that
stops at the first alarm rather than one that monitors continuously through a whole
generation.

Finally, the numbers come from one corpus and one operating regime. The transition
probabilities $p, q$, the divergence, and hence the bound are estimated on
RAGTruth, so while the qualitative ordering of detectors should transfer, the
specific values, and the $\arl$ at which recall collapses, are corpus-specific. The
delay numbers come from a trained model (seed~$42$); we quantify document-level
uncertainty with $95\%$ bootstrap confidence intervals
(Figure~\ref{fig:operating}) and additionally seed-average the decomposition over
five seeds (Appendix~\ref{app:decomp}), where the accumulation step stays robust
($3.4 \pm 0.7$ tokens) and the causal-context step stays within noise, confirming
the single-seed reading.

%% file: appendix.tex
\section{Data and Label Estimation}
\label{app:data}

We use RAGTruth \citep{niu2024ragtruth} with its official train/test split. The
test split has $2{,}700$ generations ($943$ with at least one hallucinated span,
$1{,}757$ clean) and about $341{,}000$ tokens at a $4.16\%$ hallucination rate;
the training split, on which the transition probabilities and the feature
divergence are estimated, holds about $1.98$ million tokens. A token is labeled
$1$ when it falls inside a human-annotated hallucination span and $0$ otherwise.

The transition matrix is the maximum-likelihood estimate of a two-state chain.
Every adjacent label pair $(y_{t-1}, y_t)$ over the training generations is
counted into a $2\times2$ table $C$, and each row is normalized:
\[
  p = \frac{C_{01}}{C_{00}+C_{01}}, \qquad
  q = \frac{C_{11}}{C_{10}+C_{11}}.
\]
This gives the onset hazard $p \approx 0.0044$ and the persistence
$q \approx 0.907$, a ratio $q/p > 200$. The higher-order rows of
Table~\ref{tab:markov} use the same count-and-normalize estimator with a
length-$k$ context. The labels carried by the saved detector posteriors match the
reference RAGTruth labels up to a relabeling difference of about $0.05\%$ of
tokens, too small to move $p$, $q$, or the bound.

\section{Operating-Characteristic Estimation}
\label{app:arl0}

We estimate $\arl$ on the concatenation of all hallucination-free generations.
The detector is run over this clean stream at a fixed threshold, and $\arl$ is the
total number of clean tokens divided by the number of distinct alarms, where an
alarm is counted at each upward threshold crossing and the statistic is reset
afterward. Because the stream is not segmented by document, $\arl$ is not capped
by document length, unlike a per-document false-alarm rate. Delay is measured on
the hallucination generations: the alarm position minus the onset $\theta$ when
the detector fires at or after $\theta$, and the maximum possible delay (tokens
remaining after $\theta$) when it misses or fires early, the latter giving the
censored $\edd$. Detectors are matched by sweeping their thresholds to a common
$\arl$ before any delay is compared.

\section{Speedup Decomposition and Bootstrap}
\label{app:decomp}

The decomposition of Figure~\ref{fig:operating} fixes each detector's threshold
once on the full data at the $\arl = 100$ operating point, then resamples the
hallucination generations with replacement ($B = 1{,}000$) and recomputes the
delay among detected at those fixed thresholds. This conditional bootstrap
captures document-level variance without re-matching $\arl$ on every resample.
The three steps are the nonlinear score (LogReg to HistGBM, $-12.9$ tokens, $95\%$
confidence interval $[8.8, 17.0]$), the accumulation (HistGBM threshold to CUSUM,
$-4.5$, $[1.8, 7.1]$), and the causal context (HistGBM-CUSUM to ForwardGRU-CUSUM,
$-1.9$, $[-1.0, 4.7]$); the first two intervals exclude zero and the third does
not. A ForwardGRU trained on token-shuffled sequences ($15.6$ tokens) isolates
the value of temporal order to the recurrent model itself.

The bootstrap above varies documents at a fixed model; we also seed-average the
decomposition by retraining all detectors under five seeds ($0, 1, 2, 7, 42$). The
two recurrent-model-dependent steps survive training variance: the accumulation
reduces the delay by $3.4 \pm 0.7$ tokens (robust, more than one standard deviation
from zero), the causal-context step is $0.9 \pm 1.6$ tokens (within one standard
deviation of zero), and the ForwardGRU-CUSUM delay is $12.2 \pm 1.4$ tokens. The
document-variance picture therefore carries over to training variance: the
sequential accumulation is a real effect and the extra causal context is not.

\section{Models and Features}
\label{app:models}

All detectors share the same $33$-dimensional per-token feature vector (text
statistics, NLI entailment, and generator log-probabilities) from the companion
multi-signal system. The ForwardGRU is a two-layer unidirectional GRU of hidden
width $64$ with a sigmoid output head, trained by binary cross-entropy with AdamW
(weight decay $10^{-4}$) under seed $42$ and a $15\%$ stratified validation split.
The HistGBM is a gradient-boosted classifier ($500$ trees, learning rate $0.05$,
up to $63$ leaves per tree) on the same features, and the linear baseline is a
class-balanced logistic regression. The full training recipe and the feature
extractor are released with the companion system.

\section{Closing the Gap: Details}
\label{app:gap}

The realized information rate of the learned score is
$I(\hat{g}) = \omega\,\delta_1 = 0.78$ nats per token, with Lundberg exponent
$\omega = 0.95$ and post-change drift $\delta_1 = 0.82$, against a feature
divergence $D \approx 3.5$ nats, so $D / I(\hat{g}) = 4.5$. Temperature scaling
$s \mapsto s/T$ leaves $I(\hat{g})$ exactly invariant ($\omega \mapsto \omega T$,
$\delta_1 \mapsto \delta_1 / T$); isotonic recalibration of the posterior to the
labels changes the score's shape but recovers only $+12\%$ ($0.78 \to 0.87$). On
the clean stream the score is strongly autocorrelated ($\rho_1 = 0.94$, integrated
autocorrelation time $\tau \approx 22$), and the adjustment coefficient read from
the detector's own $\arl$--threshold curve, $\omega^\star \approx 0.044$, falls
below the marginal $\omega$ by the factor $\tau$. The asymptotic dependent-data
rate then predicts a delay near $126$ tokens, an order of magnitude past the
observed $11.5$, because detection occurs well inside the score's correlation
time; this is the finite-horizon residual of Section~\ref{sec:closing}.

\section{Robustness of the Deficit}
\label{app:robust}

\paragraph{Rate anatomy.}
For a centered reference the score increment is $Y = \operatorname{logit}\hat{p} - k$
with $k = \tfrac12(\mu_0+\mu_1)$, so $\E_0[Y] = -m$ and $\E_1[Y] = +m$ with
$m = \tfrac12(\mu_1-\mu_0)$. Modeling the clean increment as $Y \sim N(-m,\sigma_0^2)$,
the Lundberg condition $\E_0[e^{\omega Y}] = \exp(-m\omega + \tfrac12\sigma_0^2\omega^2) = 1$
has the nonzero root $\omega = 2m/\sigma_0^2$, and since $\E_1[Y] = m$,
\[
  I(\hat{g}) \;=\; \omega\,\E_1[Y] \;=\; \frac{2m^2}{\sigma_0^2},
\]
which depends on the score only through the standardized separation $m/\sigma_0$ and
is therefore invariant to its scale. Table~\ref{tab:rate-anatomy} checks the form
against the empirical rate computed from the saved increments: the agreement is
within $10$--$25\%$, the increments being close to Gaussian (skewness $|{\cdot}| < 0.5$).

\begin{table}[t]
\centering
\begin{tabular}{lccccc}
\toprule
Score & $m$ & $\sigma_0$ & $m/\sigma_0$ & $I_{\mathrm{emp}}$ & $I_{\mathrm{emp}}/I_{\mathrm{gauss}}$ \\
\midrule
LogReg     & $0.38$ & $0.93$ & $0.41$ & $0.35$ & $1.04$ \\
ForwardGRU & $0.82$ & $1.25$ & $0.66$ & $0.78$ & $0.90$ \\
BiGRU      & $1.35$ & $2.13$ & $0.63$ & $1.00$ & $1.25$ \\
\bottomrule
\end{tabular}
\caption{The closed form $I = 2m^2/\sigma_0^2$ against the empirical rate. More
context raises the standardized separation $m/\sigma_0$ (LogReg to the recurrent
models), which is what lowers the deficit.}
\label{tab:rate-anatomy}
\end{table}

\paragraph{Feature augmentation (self-consistency).}
We resample the generator $K = 5$ times at temperature $0.8$ and score every token
by the mean SelfCheck-NLI contradiction of its sentence against the resamples
\citep{manakul2023selfcheckgpt}, on the Llama-2-7B-chat subset of RAGTruth. A
nearest-neighbour estimate \citep{wang2009divergence} puts the conditional
divergence this block adds at $+0.83$ nats over the $33$-dimensional law, against
$-0.59$ for a matched block of random features and $\approx 0$ under a
diagonal-Gaussian model: the gain is real and non-Gaussian. Retraining the
ForwardGRU-CUSUM on the augmented features changes nothing within five-seed noise:
token AUC $0.758{\pm}.007 \to 0.756{\pm}.006$, $I(\hat{g})$
$0.572{\pm}.029 \to 0.533{\pm}.029$, censored $\edd$
$51.3{\pm}.6 \to 52.6{\pm}1.9$ tokens (this subset's baseline differs from the
main table; the comparison is to its own baseline).

\paragraph{Rate-aware objective.}
Adding a penalty $\lambda \cdot \operatorname{Var}_0[\operatorname{logit}]$ on the
clean-token log-odds variance to the cross-entropy loss lowers $\sigma_0$ but
lowers $m$ in proportion, leaving $m/\sigma_0$ --- and the delay --- unmoved
(Table~\ref{tab:rate-aware}, five seeds, epoch chosen by held-out censored $\edd$).
A discriminant-ratio objective $(\mu_1-\mu_0)^2/\operatorname{Var}_0$ instead
collapses: it drives $\sigma_0 \to 0$ and inverts the score, so we
report the variance-penalty form. The realized rate $I$ is itself gameable this way,
which is why selection and the headline metric are the measured $\edd$, not $I$.

\begin{table}[t]
\centering
\begin{tabular}{lccccc}
\toprule
$\lambda$ & AUC & $\sigma_0$ & $m/\sigma_0$ & censored $\edd$ & recall \\
\midrule
$0$ (BCE)   & $0.785$ & $1.12$ & $0.564$ & $55.4 {\pm} 1.9$ & $0.34$ \\
$0.05$      & $0.775$ & $0.76$ & $0.565$ & $55.0 {\pm} 1.6$ & $0.34$ \\
$0.10$      & $0.772$ & $0.59$ & $0.579$ & $54.9 {\pm} 2.1$ & $0.34$ \\
$0.20$      & $0.763$ & $0.38$ & $0.584$ & $59.3 {\pm} 11.2$ & $0.29$ \\
\bottomrule
\end{tabular}
\caption{A clean-variance penalty lowers $\sigma_0$ but not the standardized
separation $m/\sigma_0$, so the censored $\edd$ does not move; $\lambda = 0.2$ begins
to collapse. The rate is scale-invariant, so a scale-only objective cannot help.}
\label{tab:rate-aware}
\end{table}

\paragraph{Multivariate accumulation.}
We replace the scalar log-odds with a multivariate statistic, in feature space and on
the recurrent hidden state. The fixed projection is LDA, $w = \Sigma^{-1}(\mu_1-\mu_0)$
with a shrinkage pre-change covariance, accumulated as a CUSUM; the adaptive one is a
window-limited generalized likelihood ratio, $\mathrm{GLR}_t = \max_{t-w \le s < t}
\lVert Z_t - Z_s\rVert^2 / (2(t-s))$ on the whitened cumulative deviations
$Z = \sum \Sigma^{-1/2}(x-\mu_0)$, a Hotelling statistic that adapts the change
direction to the data ($w = 50$). Table~\ref{tab:vector} reports the delay at
$\arl = 100$. On the raw features both are poor, since the change there is barely a
shift in the mean (Mahalanobis separation $0.40$ nats, against the $3.5$ the
diagonal-Gaussian divergence assigns to variance and shape). On the $64$-dimensional
hidden state the mean shift is real ($1.5$ nats) but the optimal linear readout only
matches the trained scalar head, and the GLR is worse: a quadratic form over $64$
dimensions raises the alarm threshold needed to hold $\arl$, costing recall.

\begin{table}[t]
\centering
\begin{tabular}{llcc}
\toprule
Space & Detector & censored $\edd$ & recall \\
\midrule
features & LDA-CUSUM            & $59.6$ & $0.34$ \\
features & GLR-CUSUM            & $80.6$ & $0.02$ \\
\midrule
hidden   & scalar logit-CUSUM  & $56.3$ & $0.31$ \\
hidden   & LDA-CUSUM            & $56.2$ & $0.31$ \\
hidden   & GLR-CUSUM            & $66.1$ & $0.22$ \\
\bottomrule
\end{tabular}
\caption{Multivariate detectors at $\arl = 100$. On the raw features the change is
not a mean shift, so a linear or quadratic accumulator gains nothing; on the hidden
state the optimal linear readout (LDA) only matches the scalar head and the Hotelling
GLR is worse. Reading the representation multivariately does not lift the deficit.}
\label{tab:vector}
\end{table}

%% file: proofs.tex
\section{Proofs}
\label{app:proofs}

Throughout, $Y_t = s(X_t) - k$ are the centered score increments, the CUSUM is
$S_t = \max(0, S_{t-1} + Y_t)$, and $\tau = \inf\{t : S_t \ge h\}$ is its stopping
time. Write $\delta_1 = \E_1[Y] > 0$ for the post-change drift and let $\omega > 0$
be the Lundberg exponent solving $\E_0[e^{\omega Y}] = 1$.

\paragraph{Proposition~\ref{prop:rate} (first-order delay).}
\emph{Delay.} After the change the increments have positive mean $\delta_1$, so
$S_t$ is a random walk with drift $\delta_1$ reflected at $0$. Write the reflected
walk as $S_t = \sum_{i \le t} Y_i + L_t$ with $L_t \ge 0$ the accumulated reflection
at the barrier. Wald's identity on the free sum gives
$\E_1[S_\tau] = \delta_1\,\E_1[\tau] + \E_1[L_\tau]$; under the post-change positive
drift the walk rarely returns to $0$, so $\E_1[L_\tau] = O(1)$ is negligible relative
to $h$. Neglecting it and the overshoot $S_\tau - h$ (bounded in expectation under a
mild integrability condition on $Y$), $\E_1[S_\tau] = h(1 + o(1))$, hence
\[
  \edd = \E_1[\tau] = \frac{h}{\delta_1}\,(1 + o(1)).
\]
\emph{False alarm.} Under $P_0$ the increments have negative mean, so the walk
drifts down and a false alarm is a large deviation. Tilt by the Lundberg exponent,
$d\tilde P_0 = e^{\omega Y}\,dP_0$; by $\E_0[e^{\omega Y}] = 1$ this is a
probability law under which the walk has positive drift, and the standard renewal
estimate for the level-crossing of a tilted walk \citep{siegmund1985sequential}
gives $\arl = \E_\infty[\tau] = e^{\omega h}(1 + o(1))$, i.e.
$h = \omega^{-1}\ln \arl\,(1 + o(1))$. \emph{Combining,}
\[
  \edd \approx \frac{h}{\delta_1}
        = \frac{\ln \arl}{\omega\,\delta_1}
        = \frac{\ln \arl}{I(s)},
  \qquad I(s) = \omega\,\delta_1. \qquad\square
\]

\paragraph{Theorem~\ref{thm:info-opt} (information-rate optimality).}
Write $D = \KL{p_1}{p_0}$. The Donsker--Varadhan variational formula states that
for every measurable $f$ with $\E_0[e^{f}] < \infty$,
\begin{equation}
\label{eq:dv}
  \E_1[f] - \log \E_0\!\big[e^{f}\big] \;\le\; D,
\end{equation}
with equality if and only if $e^{f} \propto dP_1/dP_0$ ($P_0$-a.e.). Apply
\eqref{eq:dv} with $f = \omega Y$. The Lundberg condition
$\E_0[e^{\omega Y}] = 1$ makes the second term vanish, so
\[
  I(s) = \omega\,\E_1[Y]
       = \E_1[\omega Y] - \log \E_0\!\big[e^{\omega Y}\big]
       \;\le\; D.
\]
Equality holds iff $e^{\omega Y} \propto dP_1/dP_0$, i.e.\ $\omega Y =
\log(p_1/p_0) + c$ for a constant $c$, i.e.\ $s$ is an affine function of the
log-likelihood ratio. For the delay statement, Proposition~\ref{prop:rate} gives
$\edd \approx \ln(\arl)/I(s) \ge \ln(\arl)/D$, with the lower bound attained by the
log-likelihood-ratio score, which is exactly the floor of
Theorem~\ref{thm:lorden}. $\square$

\paragraph{Remark.}
The two results compose into the gap identity used in
Section~\ref{sec:experiments}: at a false-alarm budget $\arl$, the delay of a
score $s$ is $\ln(\arl)/I(s)$ and the floor is $\ln(\arl)/D$, so the multiplicative
gap is $D/I(s) \ge 1$, the Donsker--Varadhan deficit of the score. The empirical
$D/I(\hat{g}) \approx 4.5$ for the learned score is this deficit, measured.